\documentclass[letterpaper]{article} % DO NOT CHANGE THIS
\usepackage{aaai25}  % DO NOT CHANGE THIS
\usepackage{times}  % DO NOT CHANGE THIS
\usepackage{helvet}  % DO NOT CHANGE THIS
\usepackage{courier}  % DO NOT CHANGE THIS
\usepackage[hyphens]{url}  % DO NOT CHANGE THIS
\usepackage{graphicx} % DO NOT CHANGE THIS
\usepackage{natbib}  % DO NOT CHANGE THIS AND DO NOT ADD ANY OPTIONS TO IT
\usepackage{caption} % DO NOT CHANGE THIS AND DO NOT ADD ANY OPTIONS TO IT
\usepackage{algorithm}
\usepackage{algorithmic}

\newcommand{\xmark}{\ding{55}}

\usepackage[table,xcdraw]{xcolor}
\usepackage{amsmath}
\usepackage{bm}
\usepackage{amssymb}
\newcounter{corfn}
\usepackage{booktabs} 
\usepackage{multirow} 
\usepackage{pifont}

\def\correspondingauthor{%
  \ifnum\value{corfn}=0%
    \footnote{Corresponding author.}%
    \setcounter{corfn}{\value{footnote}}%
  \else%
    \footnotemark[\value{corfn}]%
  \fi%
}%

\usepackage{newfloat}
\usepackage{listings}
\DeclareCaptionStyle{ruled}{labelfont=normalfont,labelsep=colon,strut=off} % DO NOT CHANGE THIS
\floatstyle{ruled}
\newfloat{listing}{tb}{lst}{}
\floatname{listing}{Listing}
\title{AE-NeRF: Augmenting Event-Based Neural Radiance Fields for Non-ideal Conditions and Larger Scenes}
\author{
    Chaoran Feng\textsuperscript{\rm 1},
    Wangbo Yu\textsuperscript{\rm 1,2}
    Xinhua Cheng\textsuperscript{\rm 1},
    Zhenyu Tang\textsuperscript{\rm 1},
    Junwu Zhang\textsuperscript{\rm 1},\\
    Li Yuan\textsuperscript{\rm 1,2}\correspondingauthor,
    Yonghong Tian\textsuperscript{\rm 1,2}\correspondingauthor
}
\affiliations{
    \textsuperscript{\rm 1}School of Electronic and Computer Engineering, Peking University, China\\
    \textsuperscript{\rm 2}Peng Cheng Laboratory, China \\
    \{chaoran.feng, wbyu, chengxinhua, zhenyutang, junwuzhang\}@stu.pku.edu.cn, \{yhtian, yuanli-ece\}@pku.edu.cn
}

\usepackage{bibentry}
\begin{document}

\maketitle
\newcommand{\mymodel}{\textbf{AE-NeRF}}
% -------------------- main page -----------------------------%
% \begin{abstract}
% Compared to traditional frame-based methods, computational neuromorphic imaging (CNI) using event cameras offers significant advantages, such as minimal motion blur, enhanced temporal resolution and high dynamic range. 
% The multi-view consistency of Neural Radiance Fields (NeRF), combined with the unique benefits of event cameras, has spurred recent research into reconstructing NeRF from data captured by moving event cameras. 
% While showing impressive performance, it relies on the availability of uniform and high-quality event sequences with highly accurate camera poses, thus limiting its application in real-world and unbounded scenarios.
% In this work, we propose \mymodel, to address the challenge of novel-view synthesis given non-uniform event sequences(vibrating between sparse and dense) and noisy poses.
% Our method expoilts the density of event stream and pose correction module in order to jointly learn the NeRF and refine the camera poses. 
% By relying on corrected poses, we propose hierarchical event distillation with proposal e-NeRF and vanilla e-NeRF to resample and refine reconstruction process. 
% Our event reconstruction loss and temporal loss futher encourage the reconstructed scene to be consistent from any viewpoint.
% Our approach sets a new state of the art on both synthetic and real-world challenging event datasets.
% \end{abstract}
\begin{abstract}
Compared to frame-based methods, computational neuromorphic imaging using event cameras offers significant advantages, such as minimal motion blur, enhanced temporal resolution, and high dynamic range. 
The multi-view consistency of Neural Radiance Fields combined with the unique benefits of event cameras, has spurred recent research into reconstructing NeRF from data captured by moving event cameras. 
While showing impressive performance, existing methods rely on ideal conditions with the availability of uniform and high-quality event sequences and accurate camera poses, and mainly focus on object level reconstruction, thus limiting their practical applications. 
In this work, we propose \mymodel~to address the challenges of learning event-based NeRF from non-ideal conditions, including non-uniform event sequences, noisy poses, and various scales of scenes. 
Our method exploits the density of event streams and jointly learn a pose correction module with an event-based NeRF (e-NeRF) framework for robust 3D reconstruction from inaccurate camera poses. 
To generalize to larger scenes, we propose hierarchical event distillation with a proposal e-NeRF network and a vanilla e-NeRF network to resample and refine the reconstruction process. 
We further propose an event reconstruction loss and a temporal loss to improve the view consistency of the reconstructed scene. 
We established a comprehensive benchmark that includes large-scale scenes to simulate practical non-ideal conditions, incorporating both synthetic and challenging real-world event datasets. The experimental results show that our method achieves a new state-of-the-art in event-based 3D reconstruction.
\end{abstract}
\section{Introduction}
\label{sec:1_intro}

The rapid advancement of 3D reconstruction techniques has enabled the generation of high-fidelity novel views from camera captures of a scene, further fostering numerous downstream applications, including robotics \cite{3d-aware, zhou2023palm, kerr2022evo, feng2024dit4edit,ma2024followyourpose}, 3D games \cite{xia2024video2game, condorelli2024rapid}, and scene understanding \cite{zhu2024instantswap,yu2024viewcrafter}.
However, in environments with suboptimal lighting or rapid object motion, standard RGB cameras often struggle to capture enough scene information and may experience overexposure, underexposure, or motion blur, making the captured images unsuitable for 3D scene reconstruction.
In contrast, neuromorphic sensors, like event cameras~\cite{gallego2020survey,shao2023eicil} which detect individual changes in brightness through a sequence of asynchronous events based on polarity rather than absolute intensities, provide significant advantages in such challenging situations due to their high dynamic range and temporal resolution.

\begin{figure}[t]
    \begin{center}
    \includegraphics[width=1\linewidth,trim={0.0cm 0.0cm 0.0cm 0.0cm},clip]{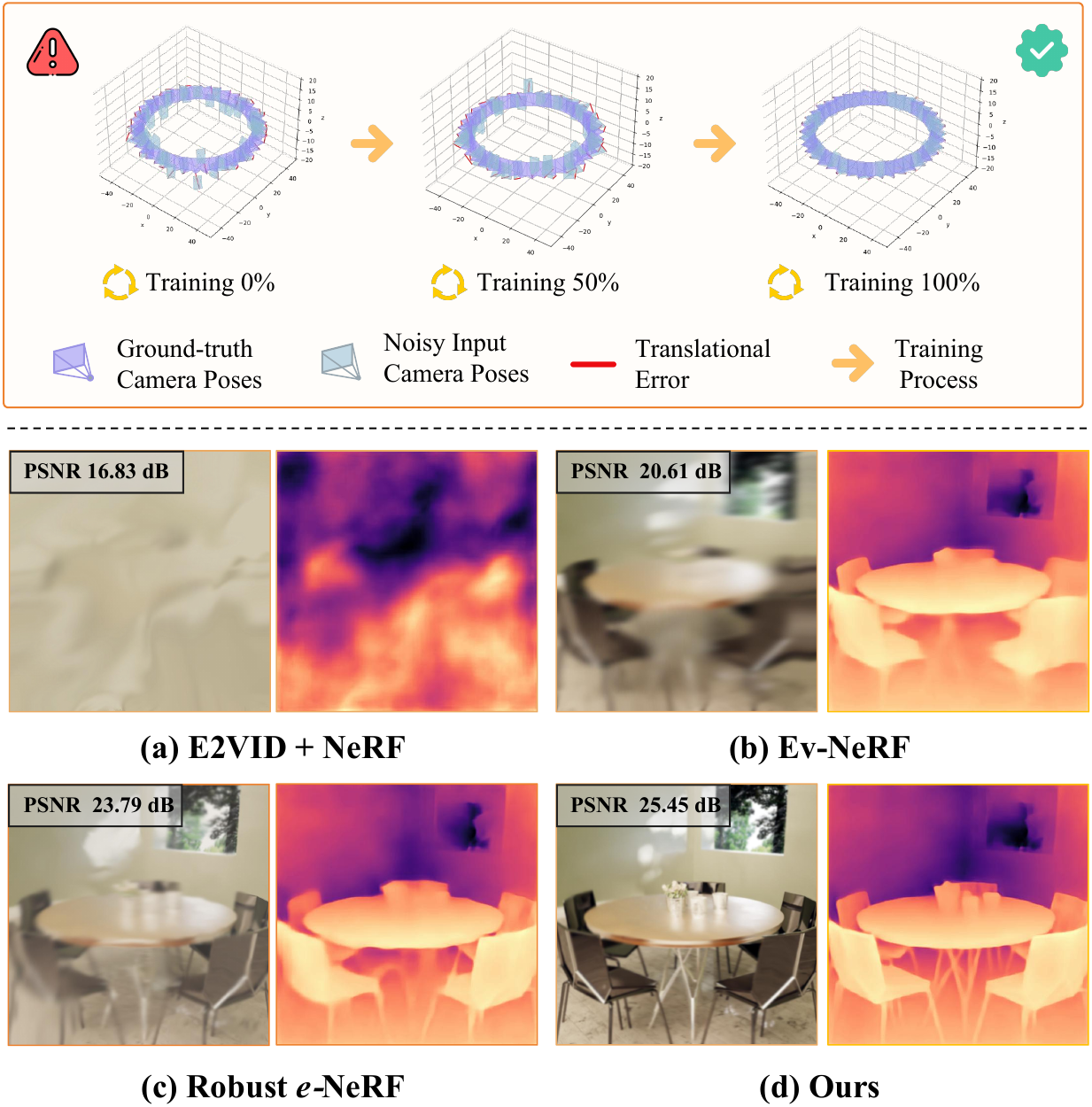}
    \end{center}
    \vspace{-0.75em}
    \caption{\textbf{Comprison of novel view synthesis (NVS) and pose correction using existing event-based methods.} The scene is captured by an event camera with 360-degree non-uniform motion and poses are estimated from COLMAP.}
    \vspace{-1.5em}
    \label{fig:comprison1}
\end{figure}
{Unfortunately, integrating event cameras into 3D reconstruction techniques remains challenging because these cameras capture relative brightness changes, which cannot be directly used to reconstruct scenes in alignment with human visual perception.}
Some methods combine depth map~\cite{li2023weakly} or standard cameras with event cameras to reconstruct 3D scenes, sacrificing the advantages of high temporal resolution offered by event cameras. Other approaches use stereo visual odometry (VO)\cite{esvo} or SLAM\cite{slam-minimal-solver-event} to address these issues, but they can only reconstruct sparse 3D models like point clouds. The sparsity limits their broader applicability. 
Additionally, another method~\cite{nehvi2021differentiable} initially represents objects as rough templates and then updates their deformations to align with events. 
However, these methods rely on template initialization, cannot address the impact of inaccurate poses, limited to specific object category scenes.

 {Neural Radiance Fields (NeRFs)~\cite{nerf} has revolutionized the field of 3D scene reconstruction by learning neural 3D scene representation from dense image captures. It also inspired Event-based NeRF reconstruction methods, such as Ev-NeRF~\cite{ev-nerf}, PAEv3d~\cite{PAEv3d}, Event-NeRF~\cite{rudnev2023eventnerf}, and Robust \textit{e}-NeRF\cite{robust-enerf}. These methods bridge NeRF with event stream (or additional RGB frame) captured by event cameras for 3D reconstruction and are desinged for handling scenes with extreme light condition or fast object motion.} 
% Moveover, it has demonstrated remarkable abilities for high-fidelity view synthesis under two conditions: \textit{dense input views} and \textit{highly accurate camera poses}. 
 {Nevertheless, these methods face fundamental challenges in non-ideal conditions that align with real-world scenarios.
Firstly, these methods rely on ground truth poses (for synthetic data) or poses derived from complicated motion capture system (for real-world data) to train NeRF. This reliance is impractical in everyday settings, where the common approach for estimating poses from captured images is to use off-the-shelf Structure-from-Motion techniques, such as COLMAP~\cite{colmap}. However, the accuracy of COLMAP rapidly degrades when handling low-quality RGB frames produced by event cameras, posing substantial challenges for the real applications of exisiting event-based NeRFs.
Secondly, non-uniform camera movement is common in real-world scenarios, which can lead to inconsistencies in the density of the event stream captured by event camera, and further affect the reconstruction qualtiy of event-based NeRFs.
Additionally, 
existing methods mainly focus on reconstructing simple objects and suffer significant performance degradation when generalized to larger scenes.
% Multiple works focus on improving event-based NeRF’s performance but struggling to address these issues.
% it compares pose correction and NVS results of existing methods
Figure \ref{fig:comprison1} illustrates an example of using existing event-based NeRFs in a large scene with non-ideal conditions. It is evident that both E2VID+NeRF~\cite{e2vid} and Ev-NeRF~\cite{ev-nerf} fail to reconstruct the 3D scene, while Robust-e-NeRF~\cite{robust-enerf} also shows low fidelity.}
% Additionally, there is a shortage of high-quality event-based challenging 3D datasets including unbounded of muti-objects scenes, as existing datasets either focus on simple objects or are generated in simulated environments with ground truth poses. This gap makes it challenging to assess model generalization in unbounded scenes, which impedes progress in related tasks, as shown in our experiments.

% The results are obtained using events generated by Blender-based designs and event simulators. 
% These above approaches of event-based NeRF obtain bad performance degradation when faced with ill-posed event data characterized by inaccurate poses, complex textures, real-world noise, and multi-object scenes. Figure \ref{fig:comprison1} compares pose estimation and reconstruction outcomes of various methods in complex indoor environments.
% Previous event-based datasets exhibit significant geometric and texture distortions caused by inaccurate poses. While methods like E2VID+NeRF and Ev-NeRF handle simple objects adequately, they struggle with color fidelity and texture consistency in complex environments and dynamic scenes. Although Robust e-NeRF improves modeling in complex motion scenarios, it still lacks robustness due to limitations in pose accuracy and the sampling strategies of the original NeRF.

In this work, to tackle the challenges of event-based NeRF reconstruction from non-ideal conditions and large scene, we make the following contributions:
\begin{itemize}
\item 
We propose \mymodel, a joint pose-NeRF training framework, which facilitates event-based NeRF reconstruction under various non-ideal conditions, particularly with inaccurate poses and uneven event density. As presented in Figure \ref{fig:comprison1}, it can effectively correct the inaccurate poses for better 3D reconstruction. 
\item 
We introduce a proposal network-based sampling strategy to address local minima optimization and large-scale generalization issues.
\item 
We propose an event-based 3D reconstruction dataset with different complex scenes based on an improved version of ESIM\cite{esim1}, setting a benchmark for event-based NeRF reconstruction in large scenes.
\end{itemize}

Comprehensive experiments validates that our method significantly outperforms prior state-of-the-art on both synthetic and real-world datasets.

% \section{Preliminary}
\begin{figure*}[h]
    \begin{center}
    \includegraphics[width=1\linewidth,trim={0.0cm 0.0cm 0.0cm 0.0cm},clip]{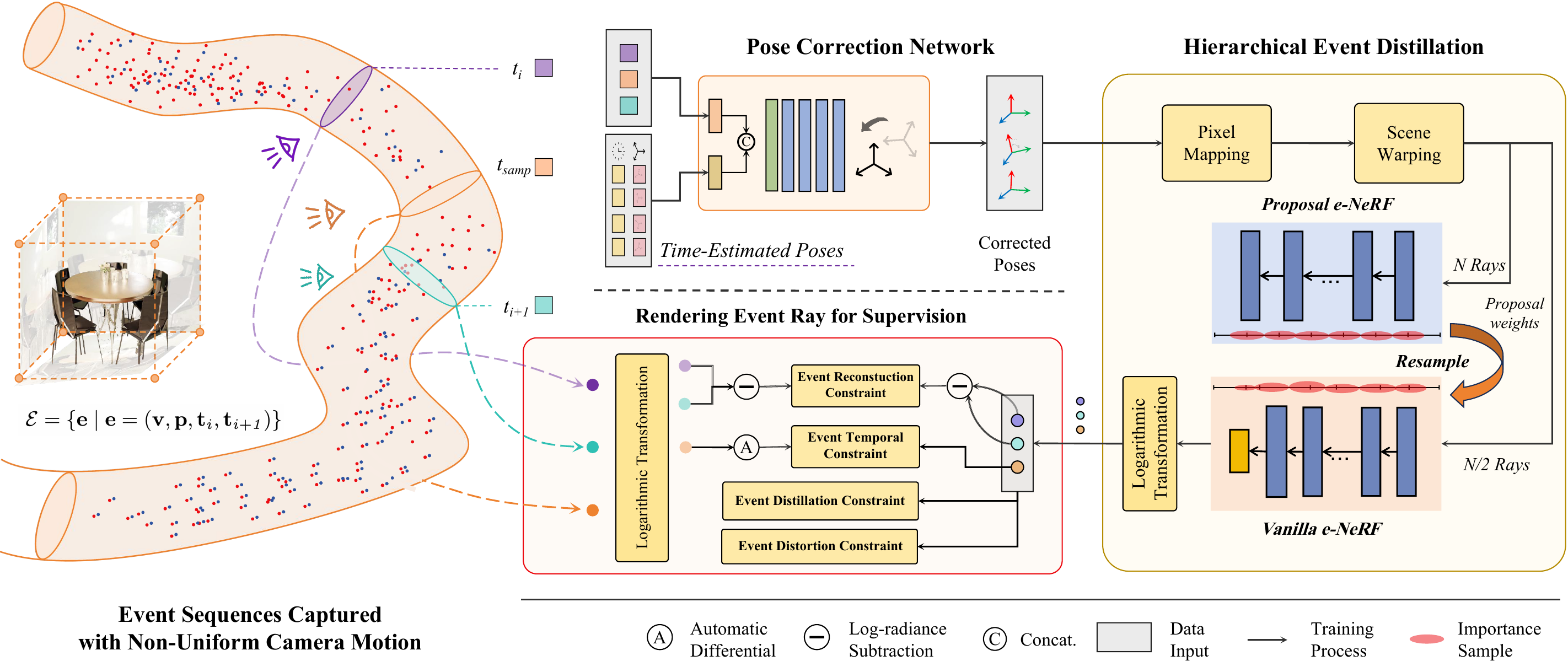}
    \end{center}
    \vspace{-0.75em}
    \caption{\textbf{Overview of \mymodel.} For each event $e$ in the batch $\mathcal{E}$, randomly sampled from the raw sequences, we sample the timestamp $t_{\text{samp}}$ between the previous timestamp $t_i$ and the current timestamp $t_{i+1}$. We then use a pose correction network with timestamps-poses pairs to interpolate discrete poses with dense timestamps, yielding corrected poses at $t_i$, $t_{i+1}$, and $t_{\text{samp}}$. With these corrected poses, we process the event ray through scene warping and apply a two-stage e-NeRF to resample weights and distances, which infers the predicted log-radiance of pixel $\mathbf{v}$. The predicted event reconstruction difference and temporal gradient are then computed against the ground truth, utilizing distillation loss and distortion loss for regularization. Finally, a learning-based approach is employed for color correction to refine tone mapping.}
    \vspace{-0.65em}
    \label{fig:pipeline}
\end{figure*}

\section{Related Work}

\subsection{3D Scene Representations}
% Previous research has explored numerous representation methods to model 3D scenes and traditional methods based on explicit representations, such as point clouds~\cite{qi2017pointnet,achlioptas2018learning_point}, mesh~\cite{wang2018pixel2mesh,liu2020general_mesh}, and voxel~\cite{byravan2023nerf2real_voxel,sitzmann2019deepvoxels}, inherently suffer from fixed topology and poor quality in novel view synthesis. 
Prior research has investigated various methods for representing 3D scenes. Traditional approaches using explicit representations, such as point clouds~\cite{qi2017pointnet,achlioptas2018learning_point}, meshes~\cite{wang2018pixel2mesh,liu2020general_mesh}, and voxels~\cite{byravan2023nerf2real_voxel,sitzmann2019deepvoxels}, often struggle with fixed topology and limited quality in novel view synthesis.
To address these issues, 3D Gaussian Splatting (3D-GS)~\cite{kerbl3Dgaussians} has been proposed, offering advantages in fast rendering speed and high-quality novel view synthesis. However, it requires point cloud initialization and does not effectively utilize the single-pixel characteristics of event stream, limiting the development of event data streams in 3D Gaussian Splatting. 

NeRF~\cite{nerf} has made significant strides by encoding continuous volumetric representations of shape and color within a multi-layer perceptron (MLP). This success has driven extensive research across computer vision, including large-scale scene reconstruction~\cite{zhang2020nerf++,mipnerf,tancik2022block-nerf,byravan2023nerf2real_voxel}, scene editing~\cite{cheng2024progressive3dprogressivelylocalediting}, scene understanding~\cite{zhi2021place-nerf-understanding}, SLAM~\cite{zhu2024nicer-slam-nerf,rosinol2023nerfslam}, and generation~\cite{tang2024cycle3d,yu2023nofa,yu2023hifi,pang2024next}. For unbounded or large-scale scenes, methods like Mip-NeRF360~\cite{barron2022mipnerf360} have enabled scene-level reconstruction, while Block-NeRF~\cite{tancik2022block-nerf} and BungeeNeRF~\cite{xiangli2022bungeenerf} have extended this to city-scale reconstruction. The integration of single-pixel event data with NeRF through pixel ray marching leverages the high dynamic range and temporal resolution of event data, ensuring high geometric consistency and texture fidelity in the reconstructed results.

\subsection{NeRF-based Reconstruction with Event Stream}
% In contrast to traditional 3D reconstruction methodologies, the domain of 3D reconstruction utilizing event cameras based on NeRF remains significantly underexplored. Contemporary NeRF-based techniques predominantly emphasize reconstruction from dense or sparse multi-view images~\cite{lu2024pano-nerf,zou2024enhancing-nerf}, potentially integrating optical flow~\cite{wang2023flow-nerf}, depth maps~\cite{depthnerf}, or point clouds~\cite{truong2023sparf-point}. The pioneering approaches for reconstructing NeRFs from event data were introduced and investigated in Ev-NeRF~\cite{ev-nerf}, E-NeRF\cite{e-nerf}, EventNeRF\cite{rudnev2023eventnerf} and Robust e-NeRF\cite{robust-enerf}. Nevertheless, these studies exhibit several limitations, as elucidated before. Moreover, EventNeRF is contingent upon accurate camera trajectory resolution, while Ev-NeRF necessitates high-quality event data streams. In scenarios characterized by imprecise pose estimation or complex scenes, these models are inadequate in maintaining geometric consistency and texture fidelity in their reconstruction outcomes.

In contrast to traditional 3D reconstruction methods, the application of event cameras in NeRF-based 3D reconstruction remains underexplored. Current NeRF techniques mainly focus on dense or sparse multi-view image reconstruction~\cite{lu2024pano-nerf,zou2024enhancing-nerf}, often incorporating optical flow~\cite{wang2023flow-nerf}, depth maps~\cite{depthnerf,li2023weakly}, or point clouds~\cite{truong2023sparf-point,jin2023diffusionret}. Early attempts to reconstruct NeRFs from event data include Ev-NeRF~\cite{ev-nerf}, E-NeRF\cite{e-nerf}, EventNeRF~\cite{rudnev2023eventnerf}, and Robust e-NeRF~\cite{robust-enerf}. However, these approaches face limitations, such as reliance on precise camera trajectories in EventNeRF and the need for high-quality event streams in Ev-NeRF. In cases of inaccurate pose estimation or complex scenes, these methodologies struggle to maintain geometric consistency and texture fidelity.

% 与传统的3D重建方法不同，基于NeRF的事件相机3D重建仍然在很大程度上未被充分探索，NeRF的衍生方法主要集中于从密集或稀疏的多视图图像进行重建，可能还包括光流、深度图或点云。最早从事件数据重建NeRFs的方法是在Ev-NeRF、E-NeRF和EventNeRF中提出和研究的。然而，这些研究存在各种限制，如第1节所述。此外，EventNeRF依赖于解析摄像机轨迹，Ev-NeRF需要较高质量的事件数据流，而在位姿估计不准确或大规模场景时，这些工作都不能较好地保证其重建结果的几何一致性和纹理一致性。

\section{Preliminary}
\subsection{Neural Radiance Fields}
\label{method:Pre:NeRF}
Our model draws its inspiration from the NeRF approach and the neural network \( F_\theta(\cdot) \) processes a 3D coordinate \( \mathbf{x}_i \in \mathbb{R}^3 \) and a ray direction \( \mathbf{d}_i \in \mathbb{S}^2 \), outputting the density \( \sigma_i \in \mathbb{R} \) and the emitted radiance \( \mathbf{c}_i \in \mathbb{R}^3 \):

\begin{equation}
F_\theta: (\gamma_x(\mathbf{x}_i), \gamma_d(\mathbf{d}_i)) \rightarrow (\sigma_i, \mathbf{c}_i),
\label{eq:nerf}
\end{equation}

Here, $\gamma(\cdot)$ is a sinusoidal positional encoding function capturing high-frequency spatial information.
Rendering each pixel involves sampling $N$ points along a ray $r(\mathbf{x}_0, \mathbf{d})$, where $\mathbf{x}_0$ is the ray's origin at the camera's focal point. The pixel color $\hat{\bm{L}}(r)$ is computed as:

% \begin{equation}
% \label{eq:L_hat}
% \hat{\bm{L}}(r) = \sum^{N}_{i=1} w_i \mathbf{c}_i,
% \end{equation}

% where \( w_i = \exp \left( -\sum^{i-1}_{l=1} \sigma_{l} \delta_{l} \right) \left( 1 - \exp \left( -\sigma_{i} \delta_{i} \right) \right) \), and \( \delta_{i} = \left\| \mathbf{x}_{i+1} - \mathbf{x}_{i} \right\| \) denotes the distance between consecutive sampling points.
\begin{equation}
\label{eq:L_hat}
\begin{aligned}
\hat{\bm{L}}(r) = \sum^{N}_{i=1} w_i \mathbf{c}_i, \delta_{i} &= \left\| \mathbf{x}_{i+1} - \mathbf{x}_{i} \right\|, \\
w_i = \exp \left( -\sum^{i-1}_{l=1} \sigma_{l} \delta_{l} \right) &\left( 1 - \exp \left( -\sigma_{i} \delta_{i} \right) \right).
\end{aligned}
\end{equation}

The ray is rendered by sampling coarse distances \( t^c \) from a uniform distribution and sorted, followed by generating coarse weights \( w^c \) with an MLP. Fine distances \( t^f \) are then sampled from the histogram with  \( t^c \) and \( w^c \), and sorted:
\begin{equation}
\label{eq:nerf_distance}
\begin{aligned}
t^c &\sim \mathcal{U}[t_n, t_f], \quad t^c = \text{sort}(\{t^c\}), \\
t^f &\sim \text{hist}(t^c, w^c), \quad t^f = \text{sort}(\{t^f\}).
\end{aligned}
\end{equation}
What's more, we adopt the normalization trick proposed in~\cite{barron2022mipnerf360} to compute the ray distance $s$.

\subsection{Event Generation Model}
\label{method:Pre:Event_Generation_Model}

% A single event is represented as \( e_k = (\mathbf{v}_k, p_k, t_k) \), indicating a brightness change detected by an event camera at time \( t_k \), with pixel location \( \mathbf{v}_k = (x_k, y_k) \) in the camera frame, and polarity \( p_k \in \{-1, +1\} \). The polarity signifies a positive or negative change in logarithmic illumination, determined by positive and negative thresholds \( C^{+1} \) and \( C^{-1} \). We follow the event generation model approach similar to that described in Robust e-NeRF \cite{robust-enerf}. An event camera captures changes in log-radiance and produces an event stream \textbf{\(\mathcal{E}\)}:

An event \( e_k = (\mathbf{v}_k, p_k, t_k) \) represents a brightness change detected by an event camera at time \( t_k \), with pixel location \( \mathbf{v}_k = (x_k, y_k) \) and polarity \( p_k \in \{-1, +1\} \). The polarity indicates positive or negative changes in logarithmic illumination, based on thresholds \( C^{+1} \) and \( C^{-1} \). We adopt the event generation model from Robust e-NeRF \cite{robust-enerf}, where an event camera captures log-radiance changes, producing an event stream \textbf{\(\mathcal{E}\)}:
\begin{equation}
    \bf{\mathcal{E}} = \{ \bm{e} \mid \bm{e} = (\bm{v}, p, t_{\mathit{i}}, t_{\mathit{i+1}}) \} \ ,
    \label{eq:event_stream}
\end{equation}
where each event records the current timestamp \( t_{\mathit{i+1}} \) and the previous timestamp \( t_{\mathit{i}} \) from the same pixel \(\bm{v}\).

An event with polarity \( p \) is triggered when the log-radiance difference at a pixel reaches the contrast threshold \( C^p \) and the condition is expressed as:
\begin{equation}
    \Delta \log \bm{L} := \log \bm{L} (\bm{v}, t_{\mathit{i+1}}) - \log \bm{L} (\bm{v}, t_{\mathit{i}}) = pC^p \ .
    \label{eq:delta_logL}
\end{equation}
For color event cameras, \( \bm{L} \) represents the radiance of light after passing through the color filter in front of the pixel.

\section{Methodology}
This work addresses the challenge of novel view synthesis based on event neural implicit representations, in the non-uniform motion and unbounded-scene regime. Especially, we assume access to only discontinuous input views with noisy camera pose estimates in real-world scenarios. 
% Our goal is to learn a neural 3D scene representation \(F_{\theta}\) from gray or colorful event streams \(\mathcal{E}\), specifically for static, complex, and unbounded scenes, even when pose estimates from COLMAP are inaccurate due to sparse and motion-blurry views in real-world scenarios.

To correct pose estimation and obtain continuous poses, we introduce a \textbf{pose correction network} \(\mathcal{\psi}(\cdot)\) using dense timestamps as the main driving signal for the joint pose-NeRF training, thereby solving the challenge of imperfect poses.
Moreover, to enhance eNeRF's ability to represent unbounded scenes, we draw inspiration from \cite{barron2022mipnerf360} and utilize \textbf{hierarchical event distillation}. This approach trains two MLPs, with one resampling and predicting volumetric density and the other handling color estimation and image rendering, and thus encourages the learned scene geometry to be consistent across all viewpoints performance.
Next, we propose and improve several normalization loss functions to \textbf{render event rays for supervision}, based on the event generation model. These functions generalize effectively to various real-world conditions, allowing joint optimization on randomly sampled event batches \(\mathcal{E}_{\mathit{batch}}\) to boost novel view rendering quality and further tackle the overfitting problem.
Finally, instead of using gamma correction, we employ a learning-based approach for \textbf{color correction} to refine tone mapping. It restores photorealistic colors from relative light intensities, enhancing overall model performance. The overall pipeline is shown in Figure~\ref{fig:pipeline}.

\subsection{Pose Correction for Continuity}
\label{method:Pose_Correction}

\begin{figure}[h]
    \begin{center}
    \includegraphics[width=1\linewidth,trim={0.0cm 0.0cm 0.0cm 0.0cm},clip]{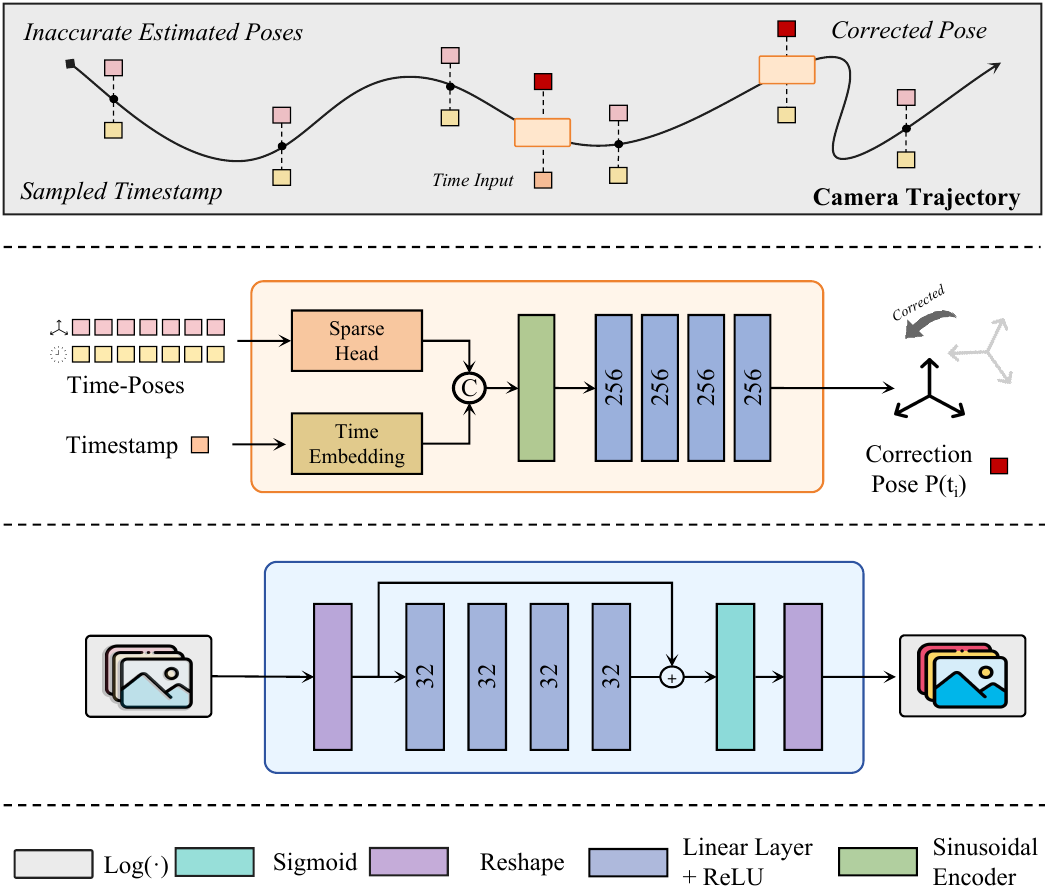}
    \end{center}
    \caption{\textbf{Framework of Pose Correction Network and Color Correction Network.}}
    \vspace{-0.85em}
    \label{fig:modules_framework_correction}
\end{figure}

Since we are accustomed to capturing RGB images and event data streams with event cameras like DAVIS346~\cite{davis}, it is inaccurate to directly estimate poses $\hat{\bm{P}_{\mathcal{E}}}$ from RGB or gray images $I_{\mathcal{E}}$ using COLMAP with fixed sampled time $T_{\mathcal{E}}$. 
The timestamp of each event $t_i \in T_{\mathcal{E}}$ and image $ I_i \in I_{\mathcal{E}}$ can form a time-image pair, serving as prior information. We achieve continuous time-pose mapping through the correction network $\mathcal{\psi}(\cdot)$. The time-image pairs are embedded via a sparse head, and each timestamp $t_i$ is also embedded. As shown in Figure~\ref{fig:modules_framework_correction} (a), this maps time to a helical axis representation $\mathcal{\psi}:=(t; T_\mathcal{E}, \hat{\bm{P}_{\mathcal{E}}}) \rightarrow SE(3)$. 

This pose correction network generates a continuous 6-DoF pose as a function of time, making it highly suitable for handling asynchronous events. Unlike other event-based NeRF approaches \cite{robust-enerf} that use trajectory interpolation or turntable poses, we address the joint problem of learning neural 3D representations and optimizing imperfect event poses, similar to BARF \cite{lin2021barf}. The process can be formulated as follows:

\begin{equation}
    \label{eq:pose_correction}
    \hat{\bm{P}}_{corr}(t_i) = \mathcal{\psi}(t_i, T_\mathcal{E}, \hat{\bm{P}_{\mathcal{E}}})
\end{equation}

where $\hat{\bm{P}}_{corr}(t_i)$ is the corrected pose at time $t_i$, $\mathcal{\psi}(\cdot)$ is the correction module. This mapping transforms the time-image pairs into a continuous $SE(3)$ pose representation, ensuring accurate pose estimationfor asynchronous event streams.

\subsection{Hierarchical Event Distillation}
\label{method:hierarchical_event_distillation}
% The multi-view correspondence loss favors a global and geometrically accurate solution, consistent across all training images. Nevertheless, the reconstructed scene often still suffers from inconsistencies when seen from novel viewpoints.
The pose correction network favors a global and continuous solution that remains consistent across all training event sequences. 
However, the reconstructed scene often exhibits inconsistencies when viewed from novel viewpoints, due to the lack of fine sampling strategies for unbounded scenes during training. 
We propose a two-phase jointly optimized e-NeRF, designed to ensure that the learned geometry is consistent from any viewing direction.
\subsubsection{Scene warping}
It has been demonstrated that space warping functions, such as NDC warping \cite{nerf} and inverse sphere warping \cite{barron2023zipnerf, f2nerf}, are effective for rendering unbounded scenes. We primarily use NDC warping for object-level scenes and employ uniform space warping $\bf{\mathcal{C(\cdot)}}$ for unbounded scenes, as defined below:
\begin{equation}
    \label{eq:space_warping}
    \bf{\mathcal{C}}(\mathbf{x}) =
    \begin{cases} 
    \mathbf{x} & \|\mathbf{x}\| \leq 1 \\
    \left(2 - \frac{1}{\|\mathbf{x}\|}\right) \left(\frac{\mathbf{x}}{\|\mathbf{x}\|}\right) & \|\mathbf{x}\| > 1
    \end{cases}
\end{equation}
\subsubsection{Two-phase Jointly Optimized e-NeRF}
We use a larger vanilla e-NeRF alongside a smaller proposal e-NeRF, repeatedly evaluating and resampling numerous samples from the proposal e-NeRF. This approach allows our model to exhibit higher capacity than existing e-NeRF methods, with only slightly increased training costs. Utilizing a small MLP to model the proposal distribution does not compromise accuracy, indicating that distilling the NeRF MLP is simpler and more effective than view synthesis, as shown below:
\begin{equation}
F_{{\varphi}}^{p}(t, \hat{w}) \circ F_{\theta}^{v}(t, w): (\gamma_x(\mathcal{C}(\mathbf{x})), \gamma_d(\mathbf{d})) \rightarrow (\sigma, \mathbf{c})
\label{eq:combined_nerf}
\end{equation}
In joint optimization process, a supervised method is needed to ensure consistency between the histograms produced by the proposal e-NeRF \(F_{\varphi}^{p}(t, \hat{w})\) and the vanilla e-NeRF \(F_{\theta}^{v}(t, w)\), as detailed in the following sections. Inspired by Mip-NeRF 360 \cite{mipnerf}, we adopt its histogram boundary function $\mathcal{B}(\cdot)$, which calculates the sum of proposal weights that overlap with interval \(T\):
\begin{equation}
    \label{eq:proposal_bound_defination}
    \mathcal{B}( \hat{t}, \hat{w}, T) = \sum_{j: T \cap \hat{T_j} \neq \emptyset} \hat{w_j}.
\end{equation}
To maintain consistency between the two histograms, for all intervals \(T_i, w_i\) within \((t, w)\), the condition \(w_i \leq \mathcal{B}(\hat{t}, \hat{w}, T_i)\) must be satisfied. This requirement resembles the additivity property of outer measures in measure theory.

This consistency ensures that the distributions between two-phase e-NeRFs remain relatively stable during resampling processes. With the boundary function, we can effectively constrain weights distribution of the proposal e-NeRF to match that of the vanilla e-NeRF, thus achieving a more efficient sampling process. It boosts the precision of sampling and the rendering quality during the training process.

% Table 1 
% Table Settings
\definecolor{lightblue}{rgb}{0.8, 0.9, 1.0}
\definecolor{lightorange}{rgb}{1.0, 0.9, 0.8}
\renewcommand{\arraystretch}{1.5}

\begin{table*}[h!]
\centering
\resizebox{0.99\textwidth}{!}{
\begin{tabular}{l@{\hskip 0.2in}ccc@{\hskip 0.2in}ccc@{\hskip 0.2in}ccc@{\hskip 0.2in}ccc}
\toprule
\multirow{2}{*}{\textbf{Methods}} & \multicolumn{3}{c}{\cellcolor{lightblue} \textbf{Easy Settings}} & \multicolumn{3}{c}{\cellcolor{lightblue} \textbf{Hard Settings}} & \multicolumn{3}{c}{\cellcolor{lightorange} \textbf{Easy Settings}} & \multicolumn{3}{c}{\cellcolor{lightorange} \textbf{Hard Settings}} \\
\cmidrule(lr){2-4} \cmidrule(lr){5-7} \cmidrule(lr){8-10} \cmidrule(lr){11-13}
& PSNR$\uparrow$ & SSIM$\uparrow$ & LPIPS$\downarrow$ & PSNR$\uparrow$ & SSIM$\uparrow$ & LPIPS$\downarrow$ & PSNR$\uparrow$ & SSIM$\uparrow$ & LPIPS$\downarrow$ & PSNR$\uparrow$ & SSIM$\uparrow$ & LPIPS$\downarrow$ \\
\midrule
E2VID+NeRF & 18.92 & .8328 & .3167 &  18.92 & .8328 & .3167 & 15.60 & .6212 & .3298 & 15.60 & .6212 & .3298 \\
Ev-NeRF & 27.72 & .9356 & .0891 & 25.44 & .8903 & .1275 & 18.26 & .7256 & .2718 & 17.82 & .6959 & .2943 \\
Event NeRF & 26.81 & .9183 & .1007 & 24.92 & .8783 & .1377 & 17.99 & .7048 & .2881 & 17.68 & .6933 & .2996 \\
Robust e-NeRF & \underline{28.38} & \underline{.9462} & .0578 & \underline{28.37} & \underline{.9464} & .0578 & 21.48 & .9033 & .1204 & 21.41 & .9028 & .1207 \\
\midrule
\textbf{Ours}(w.o. $\psi$ ) & 28.17 & .9441 & \underline{.0541} & 28.13 & .9436 & \underline{.0550} & \underline{22.83} & \underline{.9105} & \underline{.1169} & \underline{22.79} & \underline{.9098} & \underline{.1172} \\
\textbf{Ours}(w. $\psi$ )  & \textbf{28.96} & \textbf{.9512} & \textbf{.0478} & \textbf{28.90} & \textbf{.9503} & \textbf{.0485 }& \textbf{24.28} & \textbf{.9324} & \textbf{.0975} & \textbf{24.23} & \textbf{.9324} & \textbf{.0981} \\
\bottomrule
\end{tabular}
}
\vspace{-0.45em}
\caption{\textbf{Comparison of NVS for synthetic scenes.} Average performance is shown for object-level scenes in \colorbox{lightblue}{blue} and for scenes from the proposed dataset in \colorbox{lightorange}{orange}. The best and second-best results are highlighted in \textbf{bold} and \underline{underline}.}
\label{tab:synthetic_scenes}
\end{table*}

\subsection{Rendering Event Ray for Supervision}
Overfitting directly to the training event streams leads to a compromised event neural radiance field that collapses towards the provided views, even when assuming perfect camera poses. With noisy and imperfect input poses, this issue is further exacerbated, making the L2 reconstruction loss unsuitable as the primary signal for joint pose-eNeRFs training. We apply several event regularization losses, to enforce learning a globally consistent 3D solution across the optimized scene geometry and camera poses.
% As illustrated in Figure \ref{fig:pipeline}, we randomly sample a batch of events \(\mathcal{E}_{\mathit{batch}}\) from the raw, asynchronous event stream based on the event generation model, and apply several event regularization losses to supervise the performance of our model.

\subsubsection{Event Distillation Constraint.}
As mentioned earlier, we jointly optimize the proposal e-NeRF and vanilla e-NeRF, penalizing only the proposal weights that underestimate the distribution implied by the vanilla e-NeRF. Overestimation is expected since proposal weights are generally coarser and form an upper envelope over NeRF weights. This loss is similar to a half-quadratic version of the chi-squared histogram distance used in statistics and computer vision. We introduce the event threshold \(\bar{C} = \frac{1}{2} (C^{-1} + C^{+1})\) to normalize this loss, further constraining the sampling distribution. Formally, the proposal loss is defined as:
\begin{equation}
\ell_p(t, w, \hat{t}, \hat{w}) = \sum_{i} \frac{1}{w_i} \max(0, \frac{w_i - \mathcal{B}(\hat{t}, \hat{w}, T_i)}{\bar{C}})^2
\end{equation}

\subsubsection{Event Reconstuction Constraint.}
The main idea is aimed to use the log-radiance map $\Delta \log \hat{\bm{L}} := \log {\hat {\bm{L}}} (\bm{v}, t_{\mathit{i+1}}) - \log {\hat {\bm{L}}} (\bm{v}, t_{\mathit{i}})$ rendered from the training viewpoints to match ground truth relative light intensity $\Delta \log \bm{L}$. Instead of L2 Reconstruction loss of NeRF, this difference is normalized using the event threshold for supervision, as follows:

\begin{equation}
	\ell_{\mathit{r}} (\bm{e}) = \frac{\text{MSE} (\Delta \log \hat{\bm{L}}, \Delta \log {\bm{L}})}{\bar{C}^2}
    \label{eq:l_diff}
\end{equation}
Note that when using a color event camera, $\hat{\bf{L}}$ refers to the single-channel rendered radiance, where the color channel is determined by the pixel's color filter \cite{robust-enerf}.

\subsubsection{Event Temporal Constraint.}
To accurately capture the density of  event streams under non-uniform motion, we compute the error between the predicted log-radiance gradient and the target log-radiance gradient, leveraging the high sampling rate of event cameras \cite{gallego2020survey}. 
The predicted time gradient, obtained via auto-differentiation, is represented as $\mathcal{G}_{\text{pred}} = \frac{\partial}{\partial t} \log \mathbf{L}(\bm{v}, t)$, and is defined by:
\begin{equation}
\ell{\mathit{g}} (\bm{e}) = \left| \frac{\mathcal{G}_{\text{gt}} - \mathcal{G}_{\text{pred}}}{\mathcal{G}_{\text{gt}}} \right|
\label{eq:l_grad}
\end{equation}
Here, the target gradient $\mathcal{G}_{\text{gt}} \approx \frac{p C^p}{t_{\mathit{i+1}} - t_{\mathit{i}}}$ is computed as a finite difference approximation, with sampling at $t_{\mathit{sam}}$.

% To accurately capture the density of the event stream with non-uniform motion, we calculate the error using the feature of its high sampling rate between the predicted log-radiance time gradient and the target log-radiance gradient, which is derived from the temporal derivative of the event stream \cite{gallego2020survey}. The predicted time gradient is obtained through automatic differentiation, represented as $\mathcal{G}_{\text{pred}} = \frac{\partial}{\partial t} \log \mathbf{L}(\bm{v}, t)$, and is defined by:
% \begin{equation}
% \ell_{\mathit{g}} (\bm{e}) = \left| \frac{\mathcal{G}_{\text{gt}} - \mathcal{G}_{\text{pred}}}{\mathcal{G}_{\text{gt}}} \right|
% \label{eq:l_grad}
% \end{equation}
% Here, the target gradient $\mathcal{G}_{\text{gt}} := \frac{\partial}{\partial t} \log \mathbf{L}(\bm{v}, t) \approx \frac{p C^p}{t_{\mathit{i+1}} - t_{\mathit{i}}}$ is computed as a finite difference approximation, with sampling occurring at timestamp $t_{\mathit{sam}}$.

\subsubsection{Event Distortion Constraint.}
However, we notice that the depth consistency of the rendered results was not satisfactory, exhibiting some pathological depth issues. Inspired by total variance regularization~\cite{tvloss}, we adapted the smoothing from neighboring pixels to an integral over the distances between all points along the normalized ray. This integral is scaled by the weights assigned to each point by the vanilla e-NeRF, enforcing self-supervised smoothness and consistency of the ray weights:

\begin{equation}
\begin{aligned}
\ell_{d}(s, w) = & \sum_{i,j} w_i w_j \left| \frac{s_i + s_{i+1}}{2} - \frac{s_j + s_{j+1}}{2} \right| 
\end{aligned}
\label{eq:distortion_loss}
\end{equation}

In short, we use four normalized loss functions to supervise the model training with sampling a batch of events \(\mathcal{E}_{\mathit{batch}}\) and formally, the total training loss $\boldsymbol{\mathcal{L}}$ is defined as:
\begin{equation}
    \boldsymbol{\mathcal{L}} = \frac{1}{|\mathcal{E}_{\mathit{batch}}|} \sum_{\bm{e} \in \mathcal{E}_{\mathit{batch}}} ({\alpha}\ell_r+{\beta}\ell_g+{\gamma}\ell_p+{\eta}\ell_d)
    \label{eq:total_loss}
\end{equation}

\subsection{Color Correction of Synthesized Views}
\label{method:Color_Correction}
% Event cameras capture changes in log-radiance, resulting in an offset in the predicted log-radiance \(\log \hat{\bm{L}}\) from the reconstructed NeRF. In addition, a consistent proportional ambiguity exists across all color channels, similar to an image with unknown black level and ISO. These ambiguities, due to spectral sensitivity differences between event and standard cameras, can be corrected post-reconstruction using reference images. Usually, an affine correction is applied to each color channel's predicted log-radiance, or equivalently, gamma correction to the linear radiance.
% \begin{equation}
% \log \hat{\bm{L}}_\mathit{corr} = \bm{\alpha} \odot \log \hat{\bm{L}} + \bm{\beta},
% \end{equation}
% where correction parameters \(\bm{\alpha}\) and \(\bm{\beta}\) are optimally calculated via ordinary least squares using the target log-radiance \(\log \bm{L}\) from the reference image.

\begin{figure*}[!t]
    \begin{center}
    \includegraphics[width=0.94\linewidth,trim={0.0cm 0.10cm 0.0cm 0.0cm},clip]{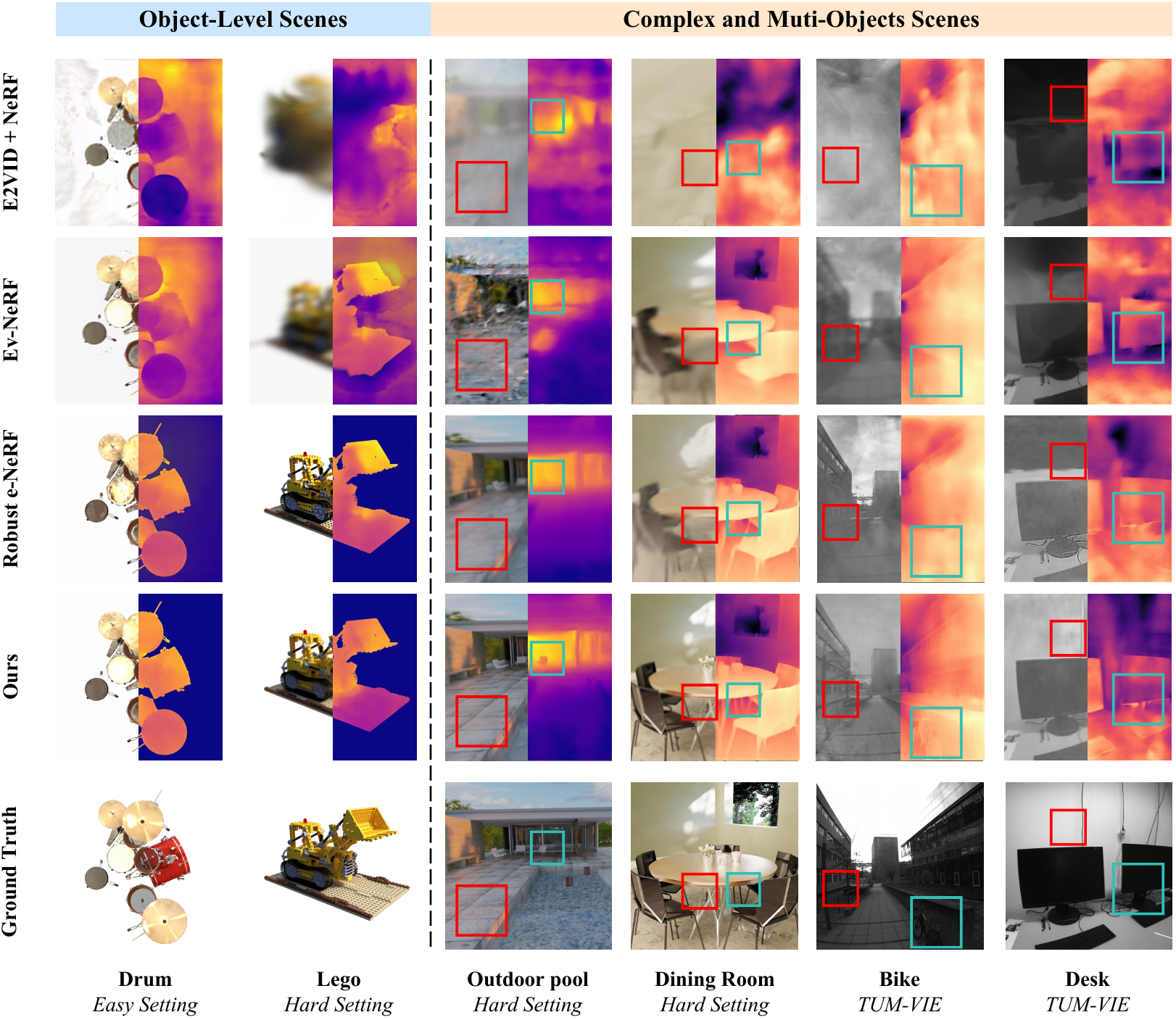}
    \end{center}
    
    \vspace{-0.75em}
    \caption{\textbf{Qualitative Comprison of Novel View Synthesis with~\mymodel.} }
    % We select six scenes from both synthetic and real-world datasets to evaluate our model with different settings, focusing on texture and geometry using RGB and depth images.
    \vspace{-0.65em}
    
    \label{fig:visulization}
\end{figure*}

Event cameras capture changes in log-radiance, leading to an offset in the predicted log-radiance \(\log \hat{\bm{L}}\) from the NeRF reconstruction. Additionally, consistent color channel ambiguities, akin to unknown black levels and ISO in images, arise from spectral sensitivity differences between event and standard cameras. These can be corrected reconstruction with affine adjustments based on reference images, with parameters optimized via ordinary least squares.

However, this method tends to perform poorly in scenes with multiple objects or complex textures environments. To further improve this process, we adopt a learning-based approach. As shown in Figure \ref{fig:modules_framework_correction}(b), we employ a color correction network $\mathcal{F(\cdot)}$ to learn the correction process for RGB images of validation views in each scene and predict the color $\hat{\textbf{c}}$ of test viewpoints from log-radiance intensity:
\begin{equation}
\label{eq:learnable_tone_mapping_eq}
\hat{\textbf{c}} = \mathcal{F}(\log \hat{\bm{L}})
\end{equation}

This approach allows for adaptive and precise tone mapping, capable of handling complex variations in the data.

\section{Experiments}
% Experiments Settings
We utilize novel view synthesis (NVS) as a benchmark to demonstrate that our method can effectively reconstruct NeRF from event cameras, particularly in scenes with inaccurate pose estimation and sparse, noisy data caused by non-uniform motion. The NVS benchmark tests are conducted on both synthetic and real sequences. In addition, we perform ablation studies on pose optimization and losses to assess the impact of each component in our method.

\textbf{Event Datasets}. For synthetic scenes, similar to Ev-NeRF and Event NeRF, we utilize the synthetic dataset from NeRF \cite{nerf} and design additional synthetic event sequences, including event data streams, estimated poses from COLMAP, ground truth poses, and RGB images, using Blender \cite{Blender}. Our dataset includes four scenes inspired by Deblur-NeRF \cite{deblurnerf} and four additional custom scenes created with Blender and ESIM. These sequences are simulated in “Realistic Synthetic 360-degree” environments, which are rich in complexity and texture, making them ideal for NVS evaluation. Following the approach of Robust e-NeRF \cite{robust-enerf}, we classify the camera trajectories into simple and challenging categories, detailed in the appendix. For real-world experiments, we use event sequences from the TUM-VIE dataset \cite{tum-vie}, which primarily capture forward motion with a Prophesee Gen 4 event camera under linear and spiral movements, providing event data, ground truth poses, and grayscale images. To simulate real-world conditions, we also estimate images using COLMAP to obtain noisy poses.

\textbf{Baseline Methods}. We compare our method with a simple baseline method E2VID+NeRF 
which combines the well-known event-to-video reconstruction method E2VID with NeRF, and recent excellent works Ev-NeRF, Event NeRF and Robust \textit{e}-NeRF.

\textbf{Evaluation Metrics}. The experimental results on both synthetic and real-world datasets are evaluated through the novel view synthesis task using quantitative metrics and qualitative comparisons of rendered images. Akin to previous studies, we use widely adopted evaluation metrics to compare synthesized images with corresponding ground truth images: PSNR, SSIM, and LPIPS to quantify the similarity between color-corrected synthesized novel views and target novel views. Moreover, we use rotation error (RE) and translation error (TE) to measure the accuracy of pose learning, further validating the model's performance.

% \section{实验}
% 我们利用新视图合成（NVS）作为基准，证明我们的方法Great e-NeRF能够直接有效地从移动事件相机中重建NeRF，特别是在位姿估计不准确以及非均匀运动产生的稀疏和噪声数据中。NVS基准测试在合成序列和真实世界序列上进行。此外，我们还对位姿优化和其他模块进行了消融研究，以评估我们方法中不同组件的影响。

% \textbf{事件数据集}. 针对合成场景，类似于Ev-NeRF和Event NeRF, 我们沿用在NeRF \cite{nerf}的合成实验数据集并基于blender\cite{Blender}设计和制作了几组合成事件序列（包括事件数据流、ground truth pose以及RGB Images），这些序列是在“真实合成360度”场景上模拟的，其拥有较为复杂的环境及纹理信息，因此对于NVS评估非常有效。受到Robost e-NeRF\cite{robust-enerf}的启发,我们也将相机轨迹分为简单和困难设置，详细情况请参照附录。真实实验是在TUM-VIE数据集\cite{tum-vie}的事件序列上进行的，这些序列主要是利用动作捕捉系统对室内进行前向捕捉，分别在线性和螺旋相机运动下利用事件相机Prophesee Gen 4得到事件相机数据、Ground Truth poses和gray images。不仅如此，我们还用Colmap对图片进行估计以得到不准确位姿，以模拟现实情况+。

% \textbf{基线方法} 我们将我们的方法Robust \textit{e}-NeRF与最近的工作Ev-NeRF \cite{ev-nerf}和一个简单的基线方法E2VID+NeRF进行对比，该基线方法通过将著名的事件到视频重建方法E2VID \cite{e2vid}与NeRF级联而成。虽然我们没有与E-NeRF \cite{e-nerf}和EventNeRF \cite{rudnev2023eventnerf}进行明确比较，但我们的实验结果仍应能准确指示它们相对于我们方法的性能，因为它们与Ev-NeRF有很多相似之处，如第X节详细描述。

% \textbf{评估指标} 我们对合成数据集和真实数据集的实验结果通过新颖的视图合成任务以三个定量指标和渲染图像之间的定性比较进行评估。与之前的研究一致，我们采用广泛使用的评估指标来将合成图像与相应的地面真实图像进行比较：\textit{峰值信噪比}（PSNR）\cite{PSNR}，\textit{结构相似性指数测量}（SSIM）\cite{ssim}，以及基于AlexNet的\textit{学习感知图像块相似性}（LPIPS）\cite{lpips}，这些指标分别评估生成图像的相对清晰度、结构相似性和感知质量，以量化伽马校正后的合成新视图与给定目标新视图之间的相似性。不仅如此，我们还使用旋转误差（RE）和平移误差(TE)用于衡量位姿学习的有效性和准确性，以更好的验证模型的性能。

% Table 2
\begin{table}[h!]
\centering
\renewcommand{\arraystretch}{1.1} 
\resizebox{0.47\textwidth}{!}{
\begin{tabular}{l@{\hskip 0.2in}c@{\hskip 0.2in}c@{\hskip 0.2in}c@{\hskip 0.2in}c@{\hskip 0.2in}c}
\toprule
\textbf{Methods} & \textbf{E2VID} & \textbf{Ev-} & \textbf{Event} & \textbf{Robust}  & \textbf{Ours} \\
& \textbf{NeRF} & \textbf{NeRF} & \textbf{NeRF} & \textbf{\textit{e}-NeRF}  & (w. $\psi$ ) \\
\midrule
PSNR$\uparrow$  & 15.81 & 17.79 & 17.61 & 18.97& \textbf{19.69}\\
SSIM$\uparrow$  & .6338 & .7856 & .7672 & .8223 & \textbf{.8470}\\
LPIPS$\downarrow$  & .3072 & .2138 & .2203  & .1964 & \textbf{.1882}\\
\bottomrule
\end{tabular}
}
\vspace{-0.5em}
\caption{\textbf{NVS Comparison the TUM-VIE dataset.}}
\label{tab:real_scenes}
\end{table}

\subsection{Evaluation on Synthetic Event Stream}
We evaluate our method on the NeRF synthetic dataset and proposed dataset comprising eight scenes. As shown in Table~\ref{tab:synthetic_scenes}, our method demonstrates notable improvement in large-scale scenes, though its impact is less pronounced in object-level scenes. It effectively preserves structural integrity, particularly at geometric discontinuities, such as the wheel in the \textit{“lego”} scene and the ground in the \textit{“outdoor pool”} scene. In contrast, the baseline method introduces significant background noise in the depth maps, whereas ours achieves clearer depth representations. Furthermore, our approach produces sharper renderings, while the benchmark results appear blurrier. Quantitative and qualitative comparisons are detailed in Table~\ref{tab:synthetic_scenes} and Figure~\ref{fig:visulization}.

\subsection{Evaluation on Real Event Stream}
We randomly select four scenes (\textit{desk1}, \textit{desk2}, \textit{office}, \textit{bike}) with different objects and materials for establishment. As stated in Figure~\ref{fig:visulization}, our approach faithfully reconstructs the main structures of objects even if there are some fog noises around them. In the scene of \textit{"bike"} , the benchmark infers wrong geometries of the bike and leads to large variances in depth predictions. Moreover, the texture and depth map of wall in the scene of\textit{ "desk" }is wrongly rendered. In contrast, ours maintains relatively clean shapes and sharper boundaries. And we additionally compute the metrics for the four scenes and the results are listed in Table~\ref{tab:real_scenes}.

\subsection{Ablations Study}
\subsubsection{Pose Correction.}
We evaluate proposed model between the ground truth poses and corrected estimated poses in outdoor pool and diningroom with hard settings, as well as office and bike in real scenes, shown in Figure~\ref{fig:comprison1} and Table~\ref{tab:ablation_pose}. It is evident that the pose correction network is significant in rectifying camera poses, particularly in complex environments where the camera motion is non-linear.
% Table 3
\begin{table}[h!]
\centering
\renewcommand{\arraystretch}{1.1} 
\resizebox{0.48\textwidth}{!}{
\begin{tabular}{l@{\hskip 0.2in}c@{\hskip 0.2in}c@{\hskip 0.2in}c@{\hskip 0.2in}c@{\hskip 0.2in}c@{\hskip 0.2in}c}
\toprule
\textbf{Scenes} & \multicolumn{2}{c}{\textbf{Synthetic Easy}} & \multicolumn{2}{c}{\textbf{Synthetic Hard}} & \multicolumn{2}{c}{\textbf{TUM-VIE}} \\
\textbf{(w. $\psi$ )} & \multicolumn{1}{c}{$\checkmark$} & \multicolumn{1}{c}{\small\xmark} & \multicolumn{1}{c}{$\checkmark$} & \multicolumn{1}{c}{\small\xmark} & \multicolumn{1}{c}{$\checkmark$} & \multicolumn{1}{c}{\small\xmark}  \\
\midrule
RE $\downarrow$ & \textbf{0.085} & 0.086 & \textbf{0.127} & 0.881 & \textbf{0.431 }& 1.283 \\
TE $\downarrow$ & \textbf{0.867} & 0.872 & \textbf{1.472} & 3.751 & \textbf{1.676}& 4.487 \\
\bottomrule
\end{tabular}
}
\vspace{-0.5em}
\caption{\textbf{Ablation study of pose correction.}}
\label{tab:ablation_pose}
\end{table}

\subsubsection{Loss Functions.}
We conclude the evaluation in the scenes same with pose correction ablation settings, in Table~\ref{tab:ablation_loss}, the contribution of all the losses introduced above. Adding event time derivative for supervision from Eq.(\ref{eq:l_grad}) improves PSNR by +3.39dB which is further increased when the event camera's trajectory is irregular. Similarly, we evaluate on adding the event distortion loss in Eq.(\ref{eq:distortion_loss}) and performance increases in this case, with a +0.57dB increase when adding this loss. The best performance is achieved when both are combined with an overall increase of +4.87dB in PSNR.

% Table 4
\begin{table}[h!]
\centering
\renewcommand{\arraystretch}{1.0} 
\resizebox{0.42\textwidth}{!}{
\begin{tabular}{c@{\hskip 0.2in}c@{\hskip 0.2in}c@{\hskip 0.2in}c@{\hskip 0.2in}c@{\hskip 0.2in}c@{\hskip 0.2in}c}
\toprule
$\ell_{p}$ & $\ell_{r}$ & $\ell_{g}$ & $\ell_{d}$ & PSNR $\uparrow$ & SSIM $\uparrow$ & LPIPS $\downarrow$ \\
\midrule
\checkmark & \checkmark & & & 19.80 & .8356 & .1811 \\
\checkmark & \checkmark & \checkmark & & 23.19 & .8852 & .1494 \\
\checkmark & \checkmark &  & \checkmark & 20.37 & .8413 & .1775 \\
\checkmark & \checkmark & \checkmark & \checkmark & \textbf{24.67} & \textbf{.8971 }& \textbf{.1337} \\
\bottomrule
\end{tabular}
}
\vspace{-0.5em}
\caption{\textbf{Ablation study of constrained loss.}}
\label{tab:ablation_loss}
\end{table}
\section{Conclusion}
This paper presents \mymodel, a novel approach designed to robustly reconstruct objects and scenes directly from moving event cameras under diverse real-world conditions. By leveraging the multi-view consistency of Neural Radiance Fieldsand the unique advantages of event cameras, our method addresses the challenges of inaccurate camera pose estimation and unbounded scene modeling through nonlinear scene parametrization, hierarchical distillation, and innovative regularizers. 
Comprehensive experiments on both synthetic and real-world datasets demonstrate that our method significantly outperforms state-of-the-art approaches and baseline benchmarks in terms of rendering quality and robustness. Our results highlight the effectiveness in enhancing novel view synthesis, even in complex environments. We will release our code and dataset to support further research and development in this field.

\section{Acknowledgments}
This work is supported in part by the Natural Science Foundation of China (No. 62332002, 62202014, 62425101, 62027804, 62088102).

\bibliography{aaai25}
\clearpage
\appendix

\section{Implementation Details}
\label{app:implementation_details}

\subsection{Datasets}
\label{app:dataset}
\subsubsection{Dataset Description}
\begin{figure*}[!ht]
    \begin{center}
    \includegraphics[width=1\linewidth,trim={0.0cm 0.0cm 0.0cm 0.0cm},clip]{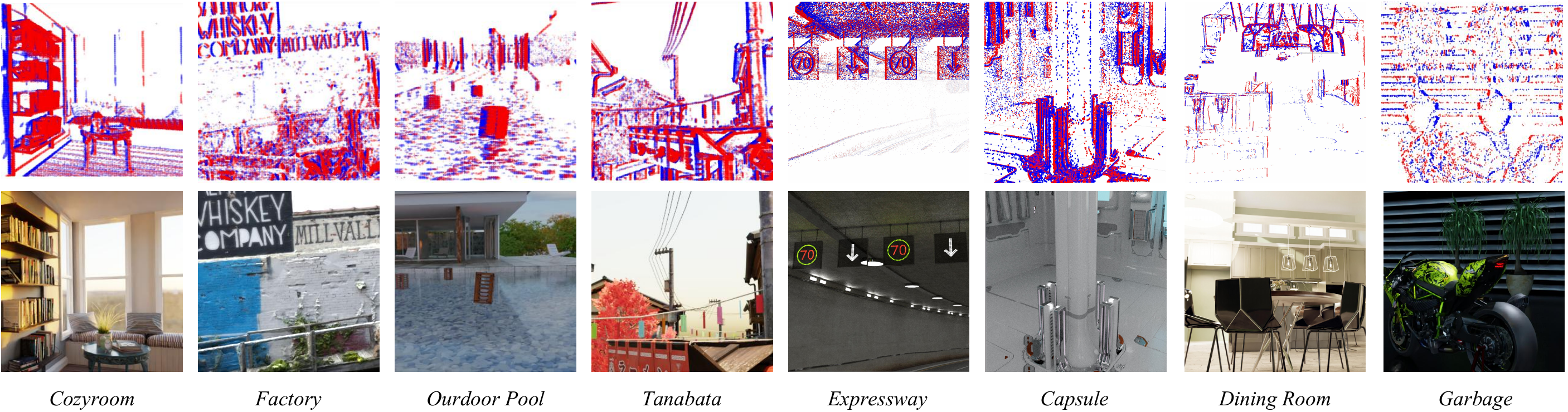}
    \end{center}
    % \vspace{-0.75em}
    \caption{\textbf{Proposed Synthetic Datasets of \mymodel.} }
    \vspace{-0.65em}
    \label{fig:app:proposed_dataset}
\end{figure*}

Existing event synthetic datasets for scene reconstruction, such as those introduced by NeRF\cite{nerf} and E2NeRF~\cite{qi2023e2nerf}, focus on object-level scenes. However, there is a notable lack of event synthetic datasets that cover large-scale scenes. Deblur-NeRF~\cite{deblur-nerf} and EvaGaussians~\cite{evagaussians} has introduced five larger-scale datasets with accompanying blender source files, but these datasets consist of rendered RGB images and do not include event sequences. To address this gap, we select four of the scenes from Deblur-NeRF and create four additional scenes ourselves, as shown in Figure~\ref{fig:app:proposed_dataset}. We generate synthetic event sequences for these scenes using ESIM~\cite{esim}, aiming to evaluate the model’s generalization ability in larger scenes.

In real-world scenarios, 3D event reconstruction methods such as PAEv3D~\cite{PAEv3d} and Ev-NeRF~\cite{ev-nerf} have been limited to object-level scenes. It is due to the constraints imposed by the need for dense and perfect pose acquisition and the resolution limits of event cameras. Datasets like TUM-VIE\cite{tum-vie} and EDS~\cite{eds}, which are used for event-based visual odometry tasks, provide perfect poses and high-quality event sequences. However, their scenes are not fully suitable for 3D reconstruction and novel view synthesis tasks. Additionally, due to the absence of a motion capture system for perfect and consistent poses, we are currently unable to capture real-world scene datasets. Consequently, we choose to quantitatively evaluate the model’s performance on real-world scenes using a subset of scenes from the TUM-VIE dataset.

\subsubsection{Dataset Settings}
We conduct experiments on both synthetic and real-world scenes to evaluate our method. For the synthetic scenes, we use the object-level dataset (\textit{chair, hotdog, materials, ficus, mic, drum, lego}) from NeRF. Due to the constraints on scene size, we design and generate several synthetic event sequences using Blender \cite{Blender} and ESIM \cite{esim}, including four datasets from the Deblur-NeRF work including \textit{outdoor pool, factory, cozyroom} and \textit{tanabata} and four additional datasets we create (\textit{Capsule, Dining Room, Garbage} and \textit{Expressway}). These sequences are simulated in "Realistic Synthetic 360-degree" scenes, which feature complex environments and texture details. As shown in Table \ref{app:tab:dataset_details}, we set the sampling rate for ground truth poses, normal images, and RGB images in Blender to 250Hz. We then use COLMAP to estimate the poses, thereby simulating real-world scenarios where camera poses are not error-free. The contrast thresholds in the ESIM simulator are set to $C^{+1} = C^{-1} = 0.25$ based on our empirical observations.

Inspired by Robust e-NeRF \cite{robust-enerf}, we divide the camera trajectories into simple and challenging settings. In the simple setting, the camera moves at a uniform speed (with RGB images potentially containing some motion blur; we use slightly motion-blurred images for COLMAP and render a separate set of non-blurred images for validation and testing). In the challenging setting, the camera speed oscillates between 1/8× and 8× of the original speed at a frequency of 1Hz. Moreover, we use 100 surrounding viewpoints as the validation set and 100 randomly select novel viewpoints as the test set to evaluate the performance of our model.

Given the difficulty of obtaining absolutely accurate poses in real-world scenarios, our real-world experiments were conduct using event sequences from the TUM-VIE dataset. These sequences were captured indoors using a motion capture system to obtain forward-facing event camera data, ground truth poses, and gray images, under both linear and spiral camera movements, with a Prophesee Gen 4 event camera. We select four scenes (\textit{mocap-desk1,mocap-desk2, office-maze, bike}) from this dataset. Additionally, we randomly select 30 novel views to evaluate the performance of our approach. Finally, although the task setting of PAEv3D dataset is not entirely aligned with our objectives, we still performd a comparison with their method. The results of this comparison are presented in the following section.

\begin{table}[h!]
\centering
\begin{tabular}{lr}
\toprule
\textbf{Settings} & \textbf{Value} \\
\midrule
Event $\mathcal{\bm{E}}$ & - \\
RGB images $I_{\mathcal{E}}$ & 250Hz \\
Normal images $N_{\mathcal{E}}$ & 250Hz \\
Ground Truth Poses $\bm{P}_{\mathcal{E}}$ & 250Hz \\
Positive Threshold of ESIM $C^{+1}$ & 0.25 \\
Negative Threshold of ESIM $C^{-1}$ & 0.25 \\
Refractory Period & 0 ns \\
\hline
Estimated Poses $\hat{\bm{P}}_{\mathcal{E}}$ & 250Hz \\
\bottomrule
\end{tabular}
\caption{\textbf{Detailed Settings of Proposed Synthetic Datasets}}
\label{app:tab:dataset_details}
\end{table}

\subsection{Model Architecture}
\mymodel~adopts Mip-NeRF360\cite{mipnerf} as the NeRF backbone due to its ability to produce high-quality reconstructions with relatively low memory consumption. Specifically, we utilize the implementation provided by the NerfAcc toolbox~\cite{li2022nerfacc}.
The parameters of the embedded Multi-Layer Perceptron (MLP) are initialized using the PyTorch-default method rather than Xavier initialization. Moreover, we replace all Rectified Linear Unit (ReLU) activations in the hidden layers with SoftPlus, as it is infinitely differentiable, which facilitates the optimization of $\ell_{g}$.

Initially, we employ the pose correction network $\psi(\cdot)$ to enhance the accuracy of the estimated poses, as outlined in Algorithm 1. The network $\psi(\cdot)$ refines time-estimated poses by integrating spatial and temporal information within a structured architecture. The network begins by processing the time-estimated poses $(\hat{\bf{P}}_\mathcal{E}, T_{\mathcal{E}})$ through a sparse head, which consists of a linear layer that outputs a 256-dimensional pose embedding, followed by SoftPlus activation. Simultaneously, the specific time instance $t_i$ is processed through another linear layer, also producing a 256-dimensional time embedding, activated by SoftPlus. These two embeddings are then combined into a single representation, which is further enhanced by a Sinusoidal Encoder that injects smooth temporal information. The enriched representation is subsequently passed through a sequence of four fully connected layers, each with 256 hidden units and SoftPlus activation, refining the feature representation. Finally, the network outputs the corrected pose $\hat{\bf{P}}_{corr}(t_i)$ via a linear transformation applied to the refined embedding. This architecture effectively integrates spatial and temporal cues, enabling precise pose correction in dynamic environments.

Then, we employ a \textit{proposal e-NeRF} with four layers and 256 hidden units for the MLP layers, and a Vanilla e-NeRF with eight layers and 1024 hidden units for the MLP layers. Both configurations use SoftPlus for internal activation and density activation. We input samples for two stages of evaluation and resampling for each \textit{proposal e-NeRF} to generate $(\hat{t},\hat{w})$, and then use half the number of samples to evaluate a single stage of \textit{vanilla e-NeRF} to generate $(t, w)$.

Additionally, we add a small $\epsilon = 0.001$ to the positive raw radiance output from the NeRF model (i.e., $\hat{\bf{\textit{L}}} = \hat{\bf{\textit{L}}} + \epsilon$) to improve the numerical stability of the predicted log-radiance $\hat{\bf{\textit{L}}}$. This augmentation imposes a lower bound of $\epsilon$ on the radiance our method can model, ensuring $\hat{\bf{\textit{L}}} > \epsilon$.

For both synthetic and real scenes, we appropriately predefine the Axis-Aligned Bounding Box (AABB), as well as the near and far bounds of the back-projected rays used for volume rendering, for each scene.

\begin{algorithm}
\label{algo:pose_corrcetion_network}
\caption{Pose Correction Network $\psi(\cdot)$}
\begin{algorithmic}[1]

    \STATE \textbf{Input:} Time-Estimated Poses ($\hat{\bf{P}}_\mathcal{E},T_{\mathcal{E}}$), Time Input $t_i$
    \STATE \textbf{Output:} Corrected Pose $\hat{\bf{P}}_{corr}(t_i)$
    
    \STATE $P_{embedding}\gets \text{ReLU}(\text{Linear}(\hat{\bf{P}}_\mathcal{E},T_{\mathcal{E}}))$
    \STATE $T_{embedding} \gets \text{ReLU}(\text{Linear}(t))$
    \STATE $C \gets P_{embedding} + T_{embedding} $
    \STATE $C \gets \text{Sinusoidal Encoder}(C)$
    
    \FOR{$i = 1$ to $4$}
        \STATE $C \gets \text{ReLU}(\text{Linear}(C))$
    \ENDFOR
    
    \STATE $\hat{\bf{P}}_{corr}(t_i) \gets \text{Linear}(C)$
    
    \STATE \textbf{Return} $\hat{\bf{P}}_{corr}(t_i)$

\end{algorithmic}
\end{algorithm}

\subsection{Training}
We implemente our method using the code frameworks of PyTorch-Lightning, Robust e-NeRF\cite{robust-enerf}, and NerfAcc\cite{li2022nerfacc}, and conducted the training on an Nvidia A6000 GPU. 
The training loss weights for all experiments are set as follows: $\lambda_{\alpha} = 1.00$, $\lambda_{\beta} = \lambda_{\eta} = 0.001$, and $\lambda_{\gamma} = 0.0025$. Following the recommendations from Instant-NGP \cite{mueller2022instant} and Robust \textit{e}-NeRF \cite{robust-enerf}, we applied a weight decay of $10^{-6}$ to the MLP to mitigate overfitting. The model underwent 50,000 training iterations, with the learning rate reduced by a factor of 0.33 at 20,000, 30,000, and 36,000 iterations (\textit{i.e.}, at 50\%, 75\%, and 90\% of the training process), as utilized in NerfAcc \cite{li2022nerfacc}. We employ the Adam optimizer with an initial learning rate of 0.01 and default hyperparameters provided by PyTorch\cite{paszke2019pytorch}. The total time of training process takes 5 hours and 30 minutes.

During the joint optimization of the contrast threshold, a higher learning rate of 0.05 is assigned to its parameter to ensure rapid convergence. The event batch size is dynamically adjusted based on the average number of ray samples required to render a single pixel, akin to the approach used in Instant-NGP, to maximize GPU memory utilization. Specifically, each batch of events contained approximately 65,536 samples in total. 

Additionally, for both the real scene target novel view poses and the synthetic scene poses without correction, we interpolate the unsynchronized constant-rate camera poses using Linear Interpolation (LERP) and Spherical Linear Interpolation (SLERP).

\section{Additional Experiment Results}

\subsection{Visualization of Ray Sampling}
\begin{figure}[!ht]
    \begin{center}
    \includegraphics[width=1\linewidth,trim={0.0cm 0.0cm 0.0cm 0.0cm},clip]{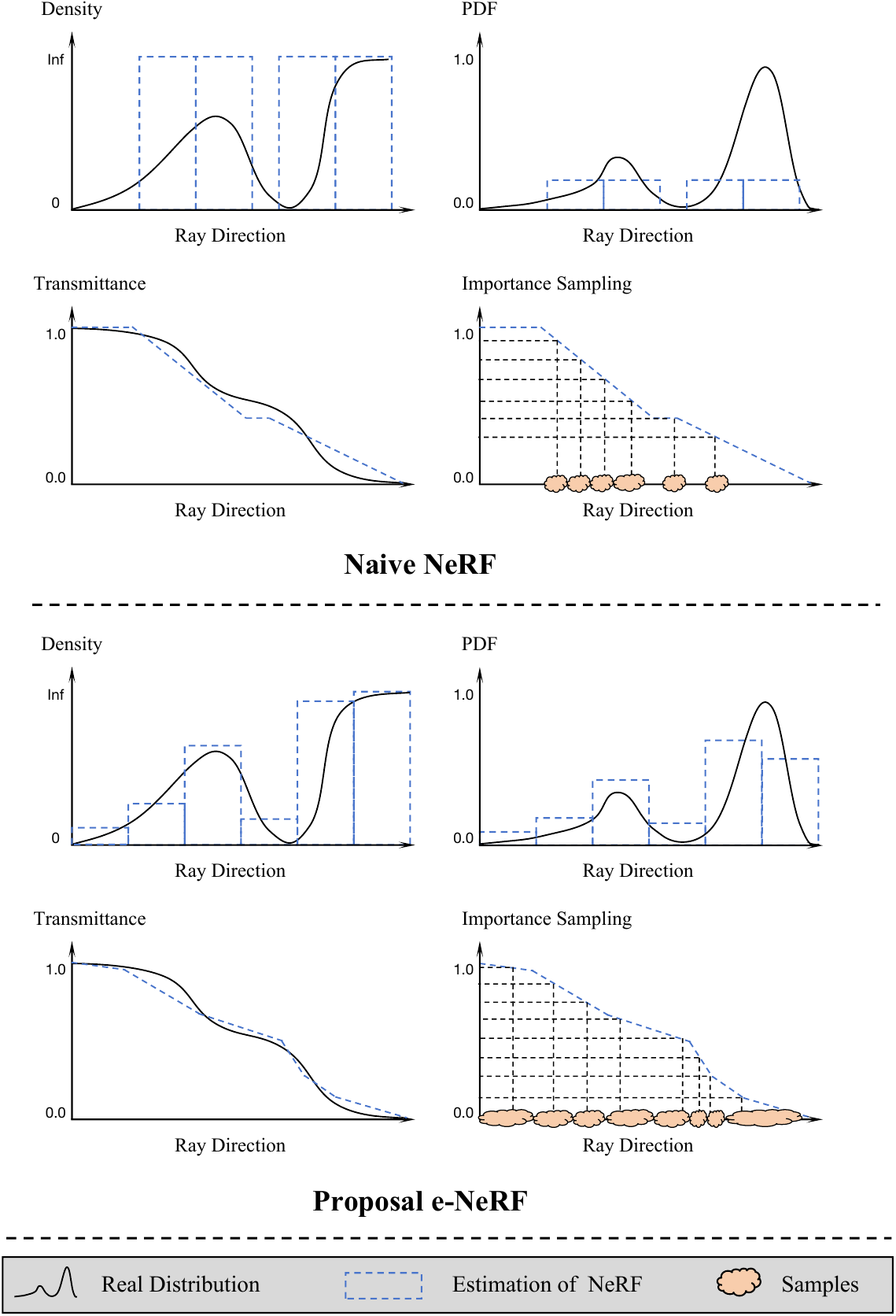}
    \end{center}
    % \vspace{-0.75em}
    \caption{\textbf{Importance Sampling Comprison.} }
    \vspace{-0.65em}
    \label{fig:app:simple_distribution}
\end{figure}
Instant-NGP leverages an occupancy grid to efficiently cache scene density using a binarized voxel grid. During ray sampling, the grid is traversed with predetermined step sizes, allowing the algorithm to bypass empty regions by querying the voxel grid. Conceptually, the binarized voxel grid serves as an estimator of the radiance field, offering significantly faster readout. Formally, this estimator represents a binarized density distribution along the ray, governed by a conservative threshold $\hat{\sigma}$ and the corresponding piecewise linear transmittance $T(t_i)$:

\begin{equation}{
\hat{\sigma}(t_i) = \mathbf{1}\left[\sigma(t_i) > \tau\right]
}
\end{equation}

As a result, the piecewise constant probability density function (PDF) can be expressed as:

\begin{equation}{
p(t_i) = \frac{\hat{\sigma}(t_i)}{\sum_{j=1}^{n} \hat{\sigma}(t_j)}
}
\end{equation}

\begin{equation}{
T(t_i) = 1 - \sum_{j=1}^{i-1} \frac{\hat{\sigma}(t_j)}{\sum_{j=1}^{n} \hat{\sigma}(t_j)}
}
\end{equation}
However, this method exhibits suboptimal performance in complex scenes due to its inadequate sampling approach. To address this limitation, we propose an adaptive sampling strategy, employing a two-phase e-NeRF model that enhances ray sampling efficiency and accelerates PDF construction.

Our method estimates the PDF along the ray using discrete samples directly. In the \textit{vanilla e-NeRF}, the coarse MLP is trained with a volumetric rendering loss to output a set of densities. This process yields a piecewise constant PDF $\sigma(t_i)$ and and a piecewise linear transmittance $T(t_i) $:

\begin{equation}{
p(t_i) = \sigma(t_i) \exp\left(-\sigma(t_i) \, dt\right)
}
\end{equation}

\begin{equation}{
T(t_i) = \exp\left(-\sum_{j=1}^{i-1} \sigma(t_j) \, dt\right)
}
\end{equation}
As illustrated in the Figure~\ref{fig:app:simple_distribution}, our \textit{proposed e-NeRF} model demonstrates superior performance compared to Instant-NGP in capturing fine-grained details. While there is a slight trade-off in overall performance, the expressiveness and consistency of our model are significantly improved.

\subsection{Quantitative Analysis of Pose Correction}

\begin{figure*}[!ht]
    \begin{center}
    \includegraphics[width=0.9\linewidth,trim={0.0cm 0.0cm 0.0cm 0.0cm},clip]{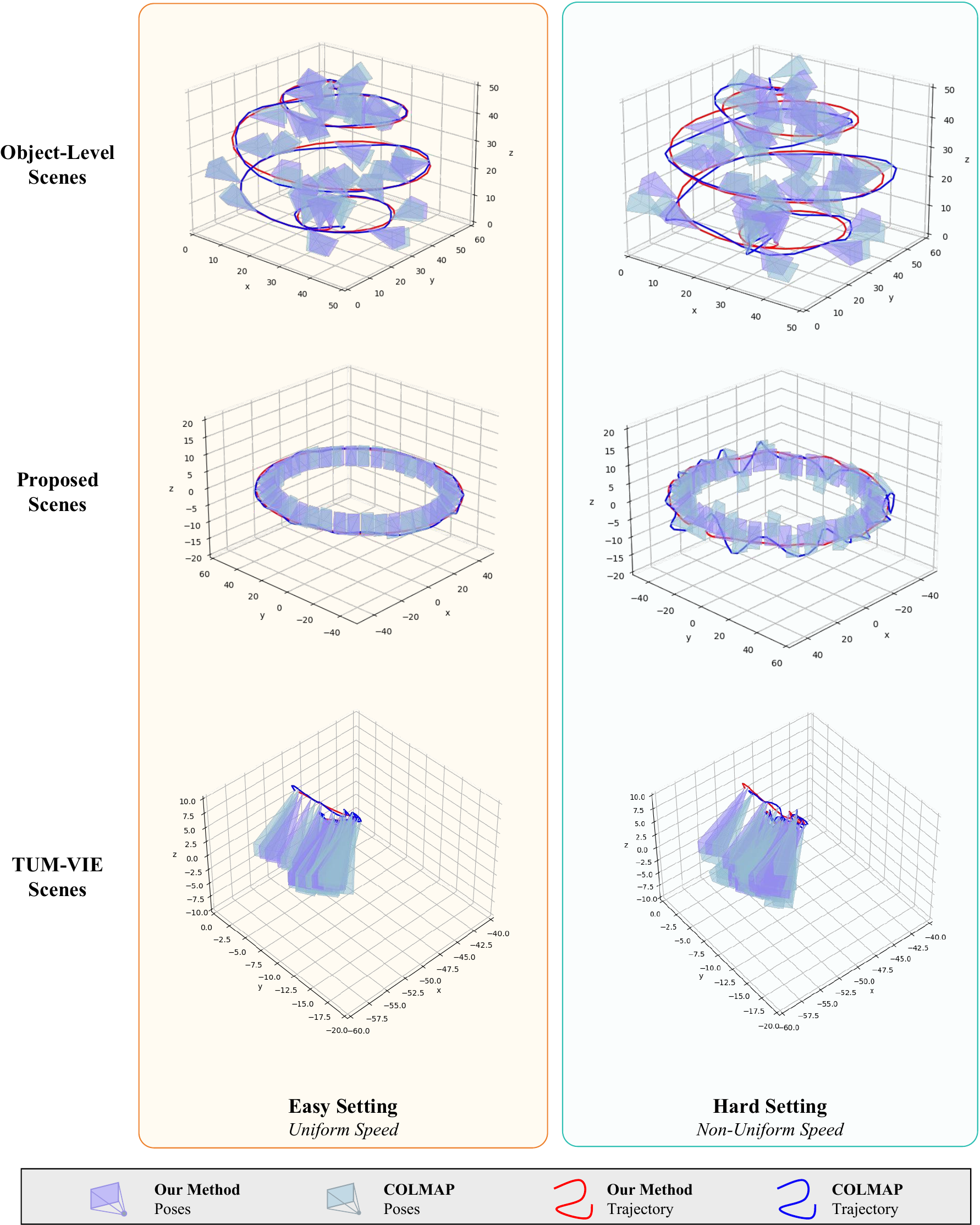}
    \end{center}
    % \vspace{-0.75em}
    \caption{\textbf{Poses Optimization Comprison.} }
    \vspace{-0.65em}
    \label{fig:app:pose_optimization}
\end{figure*}

With noisy input poses, the problem of reconstruction becomes amplified, as shown in the quantitative results and ablation study in this paper. Thus, the approach of combining dense event data to correct pose significantly improves both continuity and accuracy, as demonstrated in Figure~\ref{fig:app:pose_optimization}. In the optimization results for event sequences under \textit{uniform camera motion}, both COLMAP estimations and~\mymodel  optimizations exhibit high accuracy and consistency. However, in scenarios involving \textit{non-uniform camera motion}, COLMAP’s pose estimates become significantly inaccurate due to motion ambiguity. In contrast, our method effectively corrects these erroneous poses, leading to superior reconstruction performance.

\begin{figure*}[!ht]
    \begin{center}
    \includegraphics[width=0.9\linewidth,trim={0.0cm 0.0cm 0.0cm 0.0cm},clip]{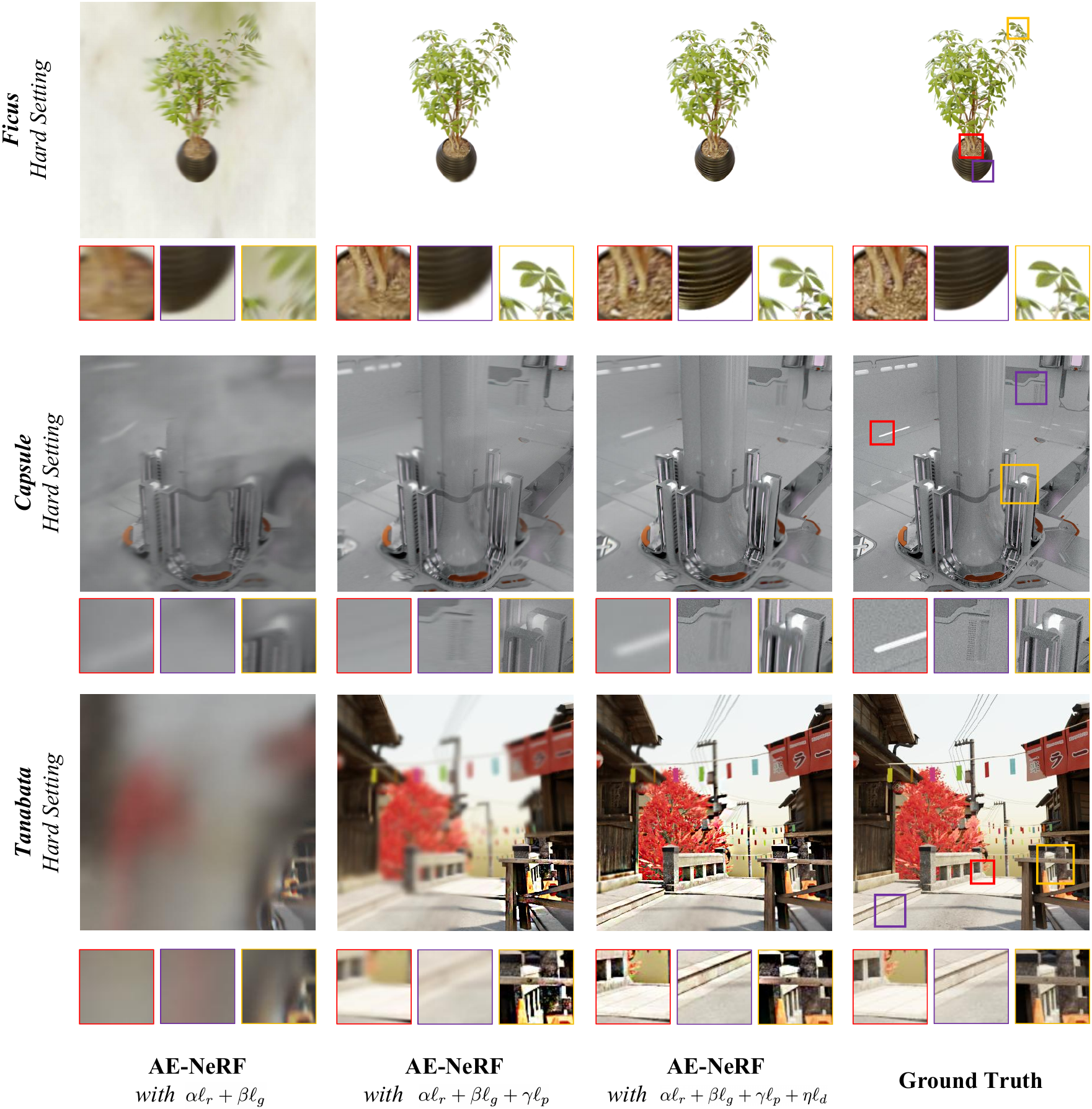}
    \end{center}
    % \vspace{-0.75em}
    \caption{\textbf{Synthesized novel views with and without losses.} }
    \vspace{-0.65em}
    \label{fig:app:loss_qualitative_analysis}
\end{figure*}

\begin{table*}[!ht]
    \centering
    \begin{tabular}{lcccc}
        \toprule
        & EventNeRF & PAEv3D & Robust e-NeRF & Ours \\
        \midrule
        PSNR & 20.273 & \underline{25.903} & 25.812 & \textbf{26.306} \\
        SSIM & .9172 & \underline{.9401} & .9377 & \textbf{.9452} \\
        LPIPS & .2278 & .0753 & \underline{.0718} & \textbf{.0715} \\
        \bottomrule
    \end{tabular}
    \caption{\textbf{Quantitative Analysis of Novel View Synthesis in the PAEv3D Datasets.} We highlight the best-performing results with \textbf{bold}, and the second-performing result with \underline{underline}.}
    \label{app:tab:paev3d_comparison}
\end{table*}

\subsection{Qualitative Analysis of Losses}
Figure~\ref{fig:app:loss_qualitative_analysis} illustrates the impact of various loss functions on the \textit{ficus} scene simulated under hard settings, the \textit{capsule} scene and the \textit{tanabata} scene sequences simulated under hard settings. It can be observed that with the inclusion of $\ell_{g}$, the \textit{ficus} scene exhibits clearer texture and geometric features. Meanwhile, the \textit{capsule} and \textit{tanabata} scenes demonstrate good reconstruction at close distances, albeit with the presence of floaters and depth inconsistencies at nearer distances. Furthermore, the results of the proposed approach combining $\ell_{g}$ and $\ell_{d}$ show a reduction in floaters and depth inconsistencies, along with sharper high-frequency details, particularly in challenging environments.

\subsection{Quantitative Analysis in PAEv3D Datasets}

In addition to the text experiments, we extend our evaluation by conducting further assessments on the dataset introduced by PAEv3D, which represents our latest advancement in event sequence-based 3Dreconstruction. For this purpose, we selected three representative scenes (\textit{bread, bounty}, and \textit{telescope}) to carry out a comprehensive quantitative analysis, as summarized in Table~\ref{app:tab:paev3d_comparison}. Although the scenarios presented in this dataset do not entirely correspond to the specific conditions and challenges of our task, our proposed method consistently demonstrates superior performance compared to existing approaches. This is particularly evident in the quantitative metrics, where our model exhibits a notable improvement. Specifically, we observe an overall increase of +0.403 dB in PSNR, underscoring the model’s enhanced capability to generalize across varied and complex environments. These results highlight the robust expressiveness of our approach, affirming its potential to effectively handle diverse real-world scenarios, even those that deviate from the original task settings.

\section{Limitation}
Due to the absence of real-world datasets featuring accurate poses, non-uniform motion, and high-quality event sequences, our current approach is limited to reconstructing the event NeRF model using synthetic event sequences under complex conditions. Future research can explore more sophisticated 3D scene event reconstruction as challenging real-world datasets become available within the community.

% \begin{table}[t!]
% \centering
% \resizebox{0.8\textwidth}{!}{
% \begin{tabular}{l@{\hskip 0.2in}ccc@{\hskip 0.2in}ccc@{\hskip 0.2in}ccc@{\hskip 0.2in}ccc@{\hskip 0.2in}ccc@{\hskip 0.2in}ccc}
% \toprule
% \textbf{Metrics} & \textbf{E2VID+NeRF} & \textbf{E-NeRF} & \textbf{Ev-NeRF} & \textbf{Event NeRF} & \textbf{Robust e-NeRF} & \textbf{Ours (w/o. PN)} & \textbf{Ours (w. PN)} \\
% \midrule
% PSNR$\uparrow$ & 30.00 & 30.00 & 30.00 & 30.00 & 30.00 & 30.00 & 30.00 \\
% SSIM$\uparrow$ & 0.132 & 0.132 & 0.132 & 0.132 & 0.132 & 0.132 & 0.132 \\
% LPIPS$\downarrow$ & 0.011 & 0.011 & 0.011 & 0.011 & 0.011 & 0.011 & 0.011 \\
% \bottomrule
% \end{tabular}
% }
% \caption{Comparison of methods for real scenes. Metrics include PSNR, SSIM, and LPIPS.}
% \label{app:tab:real_scenes}
% \end{table}

% -------------------- only supplement -----------------------------%

% \input{Sections/6_supplementary_materials}
% \bibliography{aaai25}

\end{document}